\documentclass[twocolumn, switch]{article} %

\usepackage{preprint}
\usepackage{algorithm}
\usepackage{algorithmic}
\usepackage{amsmath, amsthm, amssymb, amsfonts}

\usepackage[numbers,square]{natbib}
\usepackage[utf8]{inputenc}	%
\usepackage[T1]{fontenc}	%
\usepackage{xcolor}		%
\usepackage[colorlinks = true,
            linkcolor = purple,
            urlcolor  = blue,
            citecolor = cyan,
            anchorcolor = black]{hyperref}	%
\usepackage{booktabs} 		%
\usepackage{nicefrac}		%
\usepackage{microtype}		%
\usepackage{lineno}		%
\usepackage{float}			%

\usepackage{multirow}
\usepackage{placeins}
\usepackage{float}
\usepackage{needspace}

\usepackage{newfloat}
\DeclareFloatingEnvironment[name={Supplementary Figure}]{suppfigure}
\usepackage{sidecap}
\sidecaptionvpos{figure}{c}

\usepackage{titlesec}
\titlespacing\section{0pt}{12pt plus 3pt minus 3pt}{1pt plus 1pt minus 1pt}
\titlespacing\subsection{0pt}{10pt plus 3pt minus 3pt}{1pt plus 1pt minus 1pt}
\titlespacing\subsubsection{0pt}{8pt plus 3pt minus 3pt}{1pt plus 1pt minus 1pt}

\title{Mixed Magnification Aggregation for Generalizable Region-Level Representations in Computational Pathology}

\usepackage{xcolor}
\PassOptionsToPackage{colorlinks=true,linkcolor=gray,urlcolor=gray}{hyperref}

\usepackage{titling}
\usepackage{orcidlink}
\usepackage{footmisc}
\newcommand{\Author}[2]{%
  \textbf{#1}\textsuperscript{#2}\
}

\author{
\Author{Eric Zimmermann}{1} \and
\Author{Julian Viret}{2} \and
\Author{Michal Zelechowski}{2} \and
\Author{James Brian Hall}{1} \and
\Author{Neil Tenenholtz}{1} \and
\Author{Adam Casson}{2} \and
\Author{George Shaikovski}{2} \and
\Author{Eugene Vorontsov}{2} \and
\Author{Siqi Liu}{2} \and
\Author{Kristen A Severson}{1}
}

\date{%
  \textsuperscript{1}Microsoft Research, Cambridge, MA, United States\\
  \textsuperscript{2}Paige, NYC, NY, United States\\ 
  [0.5em]
  \footnotesize \textbf{Corresponding authors:} \texttt{ezimmermann@microsoft.com, kseverson@microsoft.com}\\
}

\begin{document}

\twocolumn[ %
  \begin{@twocolumnfalse} %

\maketitle
\thispagestyle{empty}

\begin{abstract}
In recent years, a standard computational pathology workflow has emerged where whole slide images are cropped into tiles, these tiles are processed using a foundation model, and task-specific models are built using the resulting representations. At least 15 different foundation models have been proposed, and the vast majority are trained exclusively with tiles using the 20$\times$ magnification. However, it is well known that certain histologic features can only be discerned with larger context windows and requires a pathologist to zoom in and out when analyzing a whole slide image. Furthermore, creating 224$\times$224 pixel crops at 20$\times$ leads to a large number of tiles per slide, which can be gigapixel in size. To more accurately capture multi-resolution features and investigate the possibility of reducing the number of representations per slide, we propose a region-level mixing encoder. Our approach jointly fuses image tile representations of a mixed magnification foundation model using a masked embedding modeling pretraining step. We explore a design space for pretraining the proposed mixed-magnification region aggregators and evaluate our models on transfer to biomarker prediction tasks representing various cancer types. Results demonstrate cancer dependent improvements in predictive performance, highlighting the importance of spatial context and understanding.
\end{abstract}   

\vspace{0.35cm}

  \end{@twocolumnfalse} %
] %

\section{Introduction}
\label{sec:intro}

The field of computational pathology (CPath)~\cite{deng2020deep, srinidhi2021deep, cooper2023machine, song2023artificial}, which focuses on applying machine learning models to digitized images of stained tissue samples, has rapidly adopted foundation models as a core building block for analysis. The primary samples for CPath models are whole slide images (WSIs), which are tissue samples that have been stained, affixed to glass slides, and digitally scanned. WSIs are large, often gigapixel in size, and stored as an image pyramid at various magnifications, typically 0.5, 1.0, and 2.0 microns per pixel (mpp) -- also referred to as $20\times$, $10\times$, and $5\times$ respectively. Standard computer vision approaches cannot be applied to large WSIs, therefore most CPath workflows use much smaller image crops, commonly called tiles, as the input. Foundation models use these crops, typically extracted at a single magnification of 0.5 mpp (20$\times$), along with self-supervised learning (SSL) techniques, to produce generalizable representations. 

As the CPath outcome of interest is typically associated with one or more WSIs, the numerous tile-level foundation model representations must be further aggregated. Weakly supervised models have emerged as a popular technique to achieve this second stage aggregation~\cite{campanella2019clinical,ilse18abmil,lu2021data}. Aggregation models trained using representations from foundation models  have shown great improvements in performance, e.g.~\cite{chen2024towards, vorontsov2024virchow, hoptimus0}, however, the use of foundation models does not fully address the computational challenges of CPath problems as the number of embeddings associated with a case is very large, typically ranging from one thousand to one hundred thousand~\cite{shaikovski2024prism, shaikovski2025prism2}. Furthermore, often the number of labeled cases available for supervised training is small, limiting the ability to do end-to-end fine-tuning. Some works have considered how generalizable representations at the whole-slide level might be learned, however, selecting an appropriate pretext task or supervisory signal has proven challenging ~\cite{Xu2024, ding2024multimodalslidefoundationmodel, Chen2022ScalingVT, shaikovski2024prism,shaikovski2025prism2, Wang2024, Sun2024CPathOmniAU}. Given that WSI(s) are typically associated with a pathology report, language data may serve as a natural choice and indeed language as a weak supervisory signal has demonstrated improved performance in many diagnostic tasks~\cite{shaikovski2024prism, shaikovski2025prism2, ding2024multimodalslidefoundationmodel}. However, language data may be incomplete or biased, capturing only known morphologies and lacking sufficient detail to describe low-level imaging features of interest in structural and molecular tasks. Alternatively, it is also unclear if strictly vision pretraining can provide a sufficient learning signal, as SSL algorithms often rely on whole-slide reconstruction~\cite{Xu2024} or contrastive~\cite{lenz2024unsupervised} objectives which are either lossy or fail to sufficiently compress information for perception tasks~\cite{Balestriero2024LearningBR, vanassel2025jointembeddingvsreconstruction}.

\begin{figure}
    \centering
    \includegraphics[width=0.95\linewidth]{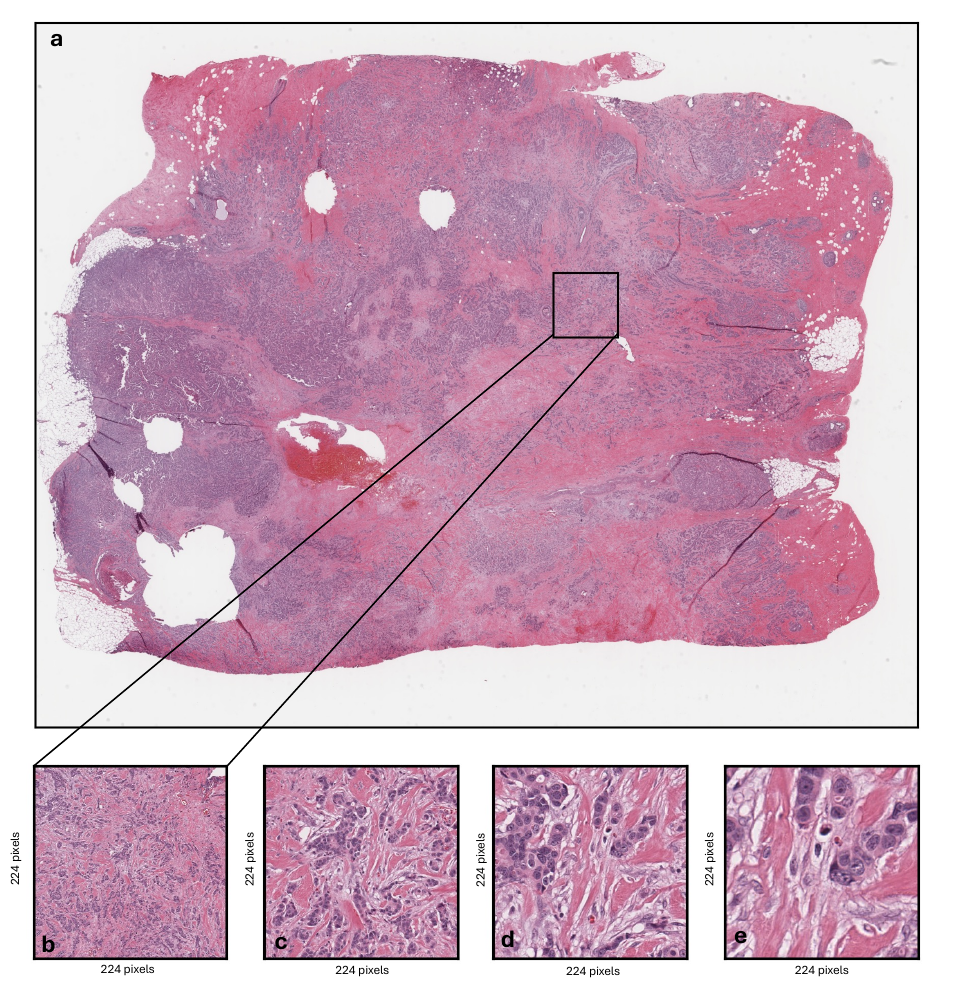}
    \caption{\textbf{a} An example of a WSI. \textbf{b} A 1344$\times$1344 micron region of the WSI. \textbf{c, d, e} Successively zoomed-in 224$\times$224 pixel regions of panel $\textbf{b}$ corresponding to 5$\times$, 10$\times$, 20$\times$, respectively. These regions depict the various features at various magnification ranging from tissue organization to individual cells.}
    \label{fig:slide_sample}
\end{figure}

Beyond computational considerations, one motivating factor for the rapid adoption of foundation models in CPath is the large variety of downstream tasks. CPath models are used for standard clinical tasks such as diagnosis, subtyping, and grading in cancer, but also are being evaluated for emerging tasks such as biomarker quantification, therapeutic response, and survival prediction. However, given this breadth of tasks, it remains an open question as to whether or not aggregating over a fixed magnification representation is sufficient. Indeed, when a pathologist reviews a WSI, they will typically zoom in and out, revealing structures and features ranging from cellular to tissue-level (see Fig.~\ref{fig:slide_sample}). Furthermore, for emerging tasks, the specific predictive morphologic features are unknown, therefore assuming a specific resolution is sufficient is undesirable. Indeed results have shown, that if predictive performance is compared across various input magnifications, a clear best magnification does not emerge~\cite{lenz2024unsupervised}.

In this work, we investigate pretraining strategies for learning generalizable region-level mixed magnification representations that can optionally compress tile-level embeddings. Given the unique setting of operating on embeddings, we explore the impact of various self-supervised learning approaches on a suite of seven downstream whole-slide biomarker prediction tasks. The two pretraining approaches of interest include masked autoencoding (MAE)~\cite{he2022masked}, which mixes tile embeddings of various magnifications extracted from Virchow2~\cite{zimmermann2024virchow}, and contrastive learning~\cite{Chen2020ASF}, which aims to compress region-level understanding by discarding redundant signals. We select Virchow2 because it is trained on four different magnifications and biomarker predictions tasks because of their variable predictive performance as a function of magnification~\cite{lenz2024unsupervised}. %
Through an exploration of the design space between tile and whole-slide modeling, we find that while self-supervised pretraining is effective for learning transferable region representations, contrastive methods do not provide a meaningful lift when incorporated into a masked embedding framework. Ultimately, our approach demonstrates that simple aggregation is often insufficient for complex biomarker prediction because magnification and spatial scales are critical factors that are not known apriori.
\section{Related works}
\label{sec:related}

\subsection{Computational pathology foundation models}
There have been many CPath foundation models proposed in recent years, starting with CTransPath~\cite{wang2022transformer}, a 15M parameter Swin Transformer model trained using approximately 15M tiles sourced from 32K publicly available WSIs. Since CTransPath's release, there have been at least 15 other foundation models proposed~\cite{ciga2022self, azizi2023robust, kang2023benchmarking, filiot2023scaling, dippel2024rudolfv, chen2024towards, Xu2024, zimmermann2024virchow, juyal2024pluto, nechaev2024hibou, hoptimus0, campanella2023computational, filiot2024phikon, wang2024pathology, aben2024towards, Xiang2025, hoptimus1, karasikov2025trainingstateoftheartpathologyfoundation}. These models vary in dataset source, e.g. public and private, dataset size, ranging from 6K to 3M WSIs, and model size, ranging from 15M to 1.9B parameters. Most use a vision transformer trained using DINOv2 and tiles at 20$\times$ magnification. Notably, Virchow2~\cite{zimmermann2024virchow}, PLUTO~\cite{juyal2024pluto}, and kaiko.ai~\cite{aben2024towards} are trained using tiles sampled at 40$\times$, 20$\times$, 10$\times$, and 5$\times$, respectively.

\subsection{Aggregation modules}
As noted in the introduction, the field of CPath has primarily converged to a two stage pipeline for whole slide inference to account for the gigapixel nature of WSI. This approach involves preprocessing WSIs by filtering for foreground tissue content and then tiling the slides using a non-overlapping grid of tissue tiles. Tile features are extracted from valid regions by leveraging foundation models. These features are then aggregated into a single slide embedding in order to make slide-level predictions, enabling accurate and scalable analysis.

Aggregation over a set of frozen tile embeddings can be done using either parametric models or by simple nonparametric reductions (e.g., mean or max). Parametric models offer more flexibility, as they allow for attentive, linear, or non-linear pooling, however, they are much more costly to train, prone to overfitting in low data regimes, and may still suffer from computational complexities due to sheer volume of embeddings that are derived from a single WSI. Models like AB-MIL~\cite{ilse18abmil}, TransMIL~\cite{Shao2021TransMILTB}, and CLAM~\cite{lu2021data} account for these complexities by attending to relevant features with learned queries, with additional clustering considerations to reduce input set size. More complex aggregators have opted to use all-in-memory transformers on entire slides~\cite{campanella2024multipleinstancelearningresolution}, while others have leveraged iterative or sparse attention~\cite{shaikovski2024prism, shaikovski2025prism2, Xu2024}, as well as subregion aggregation~\cite{ding2024multimodalslidefoundationmodel, Chen2022ScalingVT} to account for memory bottlenecks.

\subsection{Learning across multiple magnifications}

In order to capture both fine-grained cellular details and broader tissue-level context, several methods have incorporated information across magnification scales. The ability to effectively merge hierarchical concepts by capturing both micro and macro scale features may allow models to be successfully applied more generally across a variety tasks. 

Recent work has been inspired by how a pathologist may iteratively zoom in and out of a WSI to grasp multiple concepts before inferring a diagnosis. Methods like ZoomMIL~\cite{Thandiackal2022DifferentiableZF} implement a differentiable zooming aggregation operation as a means of replicating pathologist behavior. Graph neural networks have also been explored in the scope of mixed magnification aggregation, where graph structures explicitly facilitate the learning of complex long-range dependencies~\cite{Ibanez2024IntegratingMT}. Finally, HIPT (Hierarchical Image Pyramid Transformer)~\cite{Chen2022ScalingVT} presented one of the first methods to learn iteratively across magnifications with a vision transformer trained with vision-only self-supervised learning. HIPT paved the way for many future models (e.g.~\cite{shaikovski2024prism, ding2024multimodalslidefoundationmodel, Xu2024}), as it provided a successful self-supervised framework for building on top of frozen backbones. Finally, some models, such as PLUTO \cite{juyal2024pluto}, are designed for operation across magnifications, but do not explicitly mix them during inference.

While foundation models have unlocked significant performance gains, integrating aggregation strategies like multi-scale learning or graph-based fusion are still important avenues to be explored. Integration between these two components shows tremendous promise and is explored in the following section.

\section{Methods}\label{sec:methods}

We define a region mixing encoder $E_\theta$ as a parameterized transformer~\cite{Vaswani2017AttentionIA, Dosovitskiy2020AnII} with learned position encodings acting on a set of frozen foundation model tile embeddings. Specifically, the input is an ordered sequence of tile embeddings $\mathbf{x} \in \mathbb{R}^{s \times d}$, called a \textit{region embedding} of length $s$ and dimension $d$ is generated over $l$ magnification levels. The outputs of the mixing encoder are contextualized and compressed across all scales and the spatial region. We define a spatial region as a crop of the image that consists of a $t \times t$ grid of tiles at the lowest (e.g. 5$\times$) magnification. The sequence length $s$ is the count of all the tiles at all magnifications given a choice of $l$ that are contained in a spatial region . Selection of $t$ and $l$ determine the total computational cost of the mixing encoder. This selection is a compromise between sequence length, which incurs a quadratic cost in the model's attention layers, and whole slide image batch size, which is dramatically reduced with the introduction of region representations. The total sequence length $s(t,l)$ of a region embedding is:
\begin{equation}
    s(t,l) = t^2 \sum_{i=1}^l 4^{i-1}
\end{equation}
assuming that the magnifications $l$ are defined as multiples of 2, as is typical in CPath.

\begin{figure*}
    \centering
    \includegraphics[width=0.95\linewidth]{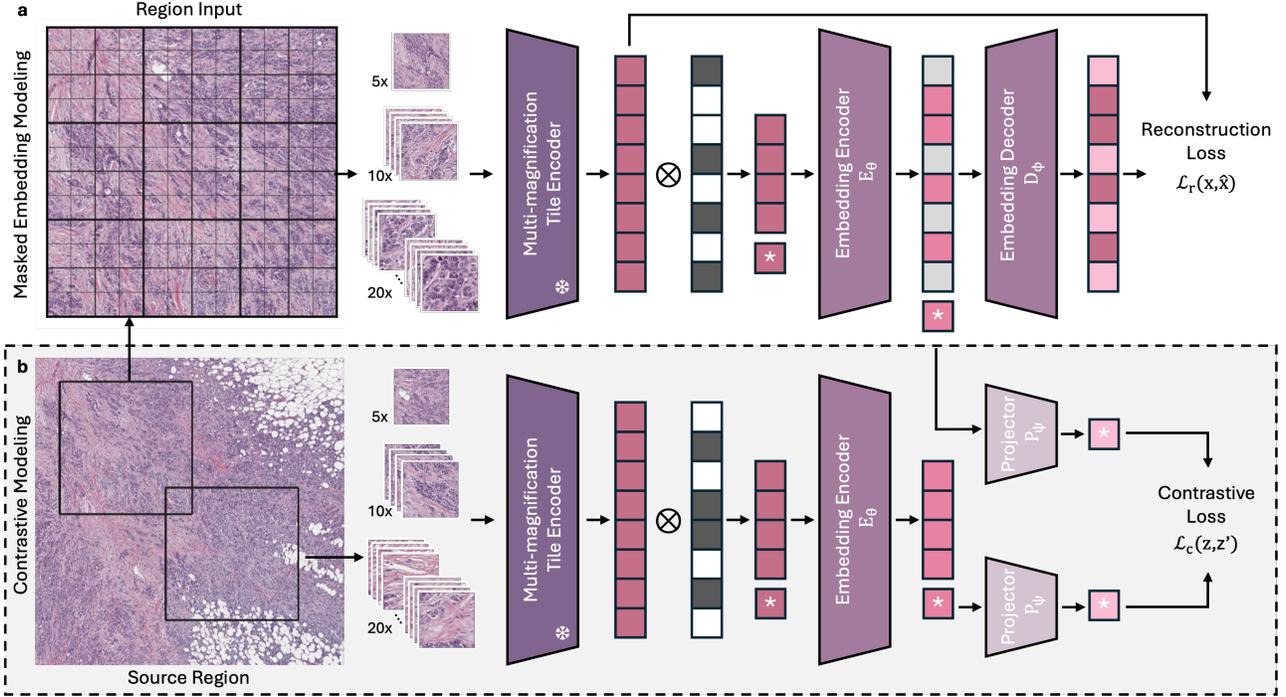}
    \caption{An overview of the pretraining framework for the mixing encoder. The basic setting takes masked embeddings from 5$\times$, 10$\times$, and 20$\times$ magnifications corresponding to a 3$\times$3 region at 5$\times$ magnification as input. \textbf{a} An encoder-decoder architecture trained using a masked reconstruction loss on patch embeddings. \textbf{b} An encoder-projector architecture using crops from a larger context as input augmentations trained with a contrastive loss on CLS embeddings denoted with *. During experimentation, two aspects of the masked reconstruction base setup are investigated: the masking rate and the addition of the contrastive branch.}
    \label{fig:method-overview}
\end{figure*}

The mixing encoder can be trained from scratch with an aggregation module for a downstream task, however, we hypothesize that the mixing encoder benefits from self-supervised pretraining. Therefore we also investigate the benefits of using highly scalable imaging based self-supervised pretraining techniques adapted to embedding based learning frameworks. Fig.~\ref{fig:method-overview} provides an overview of the approach, which is discussed in detail below.

\subsection{Self-supervised pretraining}
Self-supervised learning is a pretraining framework where targets are created from implicit knowledge in the data via a pretext task. We explore two types of pretraining using a combination of contrastive and reconstruction based objectives, operating on the prior that foundation model features are informative. 

Recent theory suggests that the choice and success of reconstruction or joint-embedding objectives are dependent on the variance of irrelevant noise present in the data and difficulty of designing augmentations, with reconstruction preferred when both are high \cite{vanassel2025jointembeddingvsreconstruction}. In practice, this guidance is hard to apply in computational pathology. Novel applications of computational pathology models such as biomarker quantification do not have well-defined morphological descriptors, can be extremely subtle, and may not be captured by a standard CPath foundation model. The lack of available training data makes it even harder to train an effective downstream aggregator. Since the features isolated during pretraining depend on both the augmentations and the learning objective, masked autoencoding is a reasonable default because it aims to preserve information content and is well positioned to capture subtle signals. However, it is not clear whether adding joint-embedding techniques provides any additional benefit, even though they are widely used for diagnostic tasks \cite{lazard2022gigasslselfsupervisedlearninggigapixel}.

\subsubsection{Masked embedding modeling}\label{sec:masked-embedding-modeling} 
Taking inspiration from the field of masked image modeling~\cite{Zhou2021iBOTIB, Peng2022BEiTVM, Chen2022ContextAF}, we investigate masked embedding modeling (MEM) in a manner similar to masked autoencoders~\cite{he2022masked}.
The MEM objective is an embedding reconstruction task using a region encoder $E_\theta$ and decoder $D_\phi$. During training, a region embedding $\mathbf{x} \in \mathbb{R}^{s \times d}$ is randomly masked at a fixed removal ratio $r \in [0, 1]$ and encoded, where the total count of removed embeddings from a sequence is $k = \lfloor s r \rfloor$.  Conditional decoding is accomplished by first inserting a learned mask token back into the encoded set at each masked location, followed by the addition of  positional encodings. The resulting decoder output embeddings are aligned with the input at each masked location using a weighted cosine similarity reconstruction loss (see Fig.~\ref{fig:method-overview}a). If \(\mathbf{m} \subset \mathbb{Z}_{s}\) is the set of indices corresponding to the \(k\) masked tokens, the sample-wise objective for a region embedding, with sequence weighting \(\mathbf{w} \in \mathbb{R}^{s}\), is
\begin{equation}
   \mathcal{L}_\textrm{r} = -\frac{1}{k} \sum_{i \in \mathbf{m}} \mathbf{w}_{i}\text{sim}\left(\hat{\mathbf{x}}_{i},\mathbf{x}_{i}\right)
\end{equation}
\noindent
where the cosine similarity between two vectors $\mathbf{u}, \mathbf{v}$ is
\begin{equation}
    \text{sim}(\mathbf{u}, \mathbf{v}) = \frac{\mathbf{u}^\top \mathbf{v}}{\Vert \mathbf{u} \Vert \Vert \mathbf{v} \Vert}
\end{equation}
\noindent
and the masked reconstructed signal $\hat{\mathbf{x}}$ is
\begin{equation}
    \hat{\mathbf{x}} = D_\phi(E_\theta(\mathbf{x} \mid \mathbf{m}) \mid \mathbf{m})
\end{equation}
Given the exponential relationship between the number of elements at a specific magnification in a region embedding, we select a set of weights such that each magnification is weighted uniformly in expectation, rather than being uniform over input embeddings.

\subsubsection{Contrastive alignment}\label{sec:contrastive-alignment}

Contrastive methods aim to learn salient representations by maximizing the similarity between positive paired samples with known correspondences while minimizing the similarity between  random unpaired negative samples~\cite{Chen2020ASF, Yeh2021DecoupledCL, HaoChen2021ProvableGF}. Sample pairing is designed via a pretext task, where a source sample is perturbed via random augmentation, while unpaired samples are considered to be random elements drawn from the data distribution. Augmentations for images are often defined as either photometric or geometric~\cite{NEURIPS2024_d2964af7,zimmermann2024virchow}, however, when dealing with embeddings, defining augmentations becomes considerably more difficult, as the input signal cannot intuitively be perturbed or inspected for correctness. 

There are two geometric augmentations that stand out as well suited to the problem when considering the input format of an embedding mixing encoder. The first augmentation is the random mask strategy, which is already used when considering the masked embedding modeling task~\cite{Xu2024,ding2024multimodalslidefoundationmodel}. The second augmentation is the random region subsample, akin to the extended-context-translation proposed in Virchow2~\cite{zimmermann2024virchow} which is similar to the general strategy of cropping without resizing~\cite{NEURIPS2024_d2964af7}. A straightforward extension of this approach to the masked embedding modeling case uses a source $c\times c$ region of embeddings with larger context and samples two random $t \times t$ target regions to be aligned by the mixing encoder (see Fig.~\ref{fig:method-overview}b).  The random region subsample aims to identify features co-occurring across nearby spatial neighborhoods. The selection of negatives combined with the distance between subregions encourages the suppression of common features present across the dataset while partially maintaining common features that are spatially consistent within a slide. Despite contrastive methods having success on both tile and whole slide level representation learning, it is unclear whether they are sufficiently well posed in the region setting, or if they could aid in learning predictive biomarker features benefit from theses augmentations. As a result, we explore this axis of variation as a means of providing a complete understanding of the design space as to inform their own design choices.

We apply contrastive learning to an extended variation of the transformer class embedding, which we call the compressed region embedding, by concatenating a series of registers~\cite{Darcet2023VisionTN} as a means of artificially increasing model capacity at a fixed model width. Compressed embedding are further processed with a lightweight feed-forward projector $P_\psi$ and are aligned using the normalized temperature-scaled cross-entropy~\cite{Chen2020ASF}. For a batch of $n$ augmented embedding output pairs $(\mathbf{z}, \mathbf{z}^\prime)$ with temperature $\tau$, the batch-wise contrastive loss indexed along the batch dimension is 
\begin{equation}
    \mathcal{L}_\textrm{c} =- \frac{1}{n} \sum_{i=1}^n\log \frac{\exp \big(\frac{1}{\tau}\text{sim}(\mathbf{z}_i, \mathbf{z}^\prime_i)\big)}{\sum^n_{j \neq i}\exp \big(\frac{1}{\tau}\text{sim}(\mathbf{z}_i, \mathbf{z}^\prime_j)\big)}
\end{equation}
\noindent
where outputs $\mathbf{z}$ are generated from a compressed embedding with
\begin{equation}
    \mathbf{z} = P_\psi(E_\theta(\mathbf{x} \mid \mathbf{m}))
\end{equation}
Note that the projector is a pretraining artifact and is discarded during inference.
 
We refer to the combination of a contrastive masked autoencoder~\cite{Huang2022ContrastiveMA} using the sum of both loss terms as CMEM. This approach has been studied in the imaging domain~\cite{Huang2022ContrastiveMA}, however, we aim to study task design in aggregation-like settings for CPath where finetuning is feasible. 

\subsection{Supervised Aggregation}

The output representations of the mixing encoder are either defined as the grouping of all patch token outputs of the transformer, which we refer to as contextualized region embeddings, or by concatenating the class tokens, which we refer to as the compressed tile embedding. Both contextualized and compressed tokens represent region-magnification fusions over region embedding inputs with different downstream capacity. These embeddings still require further aggregation over WSI(s) to a single output representation which is accomplished using an aggregator via an attention-based multiple instance learning (AB-MIL) layer~\cite{ilse18abmil}, although any aggregation technique can be used as a drop in replacement. The gated attention mechanism in AB-MIL aggregates and combines a bag encoded embeddings of set size $b$ through learned linear combination $\mathbf{h} \in \mathbb{R}^{f}$ between queries $\mathbf{V} \in \mathbb{R}^{f \times d}$ and $\mathbf{U} \in \mathbb{R}^{f \times d}$ to produce an aggregated embedding $\mathbf{z}_\textrm{agg}$ as a weighted linear combination of inputs. The output embedding are then classified using a single linear classifier. With learned parameters, the aggregation attention mechanism is described as
\begin{equation}
    \mathbf{z}_\textrm{agg} = \sum^b_{i=1} \alpha_i \mathbf{z}_i
\end{equation}
\noindent 
where the attention weights are computed as 
\begin{equation}
    \alpha_i = \frac{\exp\big(\mathbf{h}^\top (\tanh(\mathbf{V}\mathbf{z}^\top_i) \odot \sigma(\mathbf{U}\mathbf{z}^\top_i))\big)}{\sum^l_{j=1}\exp\big(\mathbf{h}^\top (\tanh(\mathbf{V}\mathbf{z}^\top_j) \odot \sigma(\mathbf{U}\mathbf{z}^\top_j))\big)}.
\end{equation}
A patient label $\mathbf{y}$ is assigned at a specimen level, where a specimen $\mathbf{G}$ is a group of at least one WSI $\mathbf{X}$. This label is sparse, and the relevant WSI which produce the label are unknown. As a result, for a binary classification task, we may further minimize resource consumption by searching for the slide with the most salient features. This is accomplished by propagating the supervisory signal to the slide that yields minimum loss $\mathcal{L}$ as follows
\begin{equation}
    \mathcal{L}(\mathbf{G}, \mathbf{y}) = \min_i \mathcal{L}(\mathbf{X}_i, \mathbf{y})
    \label{eqn:s2g}
\end{equation}

\section{Experimental Methods}
For a fixed target region size $t = 3$, we investigate both the impact of various removal ratios, $r \in \{0.25, 0.5, 0.75\}$, as well as the effect of the source region context $c \in \{3, 7,11\}$ at $r=0.5$ to assess the value of self-supervised pretraining on the mixing encoder. Further details of the experiments are explained below. 

\subsection{Mixing encoder pretraining data}
The pretraining dataset is composed of $1\textrm{M}$ hematoxylin and eosin (H\&E)  stained WSI sourced from Memorial Sloan Kettering Cancer Center (MSKCC) and includes both internally prepared and externally submitted slides across more than 200 recorded tissue types. 

Whole slide images are preprocessed by first separating foreground tissue from background glass via segmentation with a trained fully-convolutional network. Segmentation outputs are further refined with Otsu thresholding, using thresholds of $(0.4, 0.5)$ and acceptance inclusion criteria of $45\%$. Region tissue tiles of size $224 \times 224$ pixels are extracted jointly over $l=3$ levels of $5\times$, $10\times$, and $20\times$, at a target spatial resolution $t = 3$ using coordinates sampled from the segmentation map, for a total of $s(3,3) = 189$ tiles covering $2688 \times 2688$ pixels measured at $20\times$ magnification. All coordinates are grounded to a common level such that proper spatial alignment between each magnification in a region is preserved. Region tiles are encoded using Virchow2~\cite{zimmermann2024virchow}, a mixed magnification foundation model that is pretrained on billions of tiles at several magnifications and across many tissue types. Only the class token from Virchow2 is used and is of dimension $1280$.

\subsection{Mixing encoder architecture and pretraining}
A variation of the vision transformer with four class tokens, akin to registers~\cite{Darcet2023VisionTN}, is used as the mixing encoder and decoder, where the input convolutional layer is replaced with a linear layer. For all settings, the encoder has a hidden dimension of size $1536$ and is composed of $12$ layers with $12$ heads and a multi-layer perceptron (MLP) ratio of $2$ using the SwiGLU activation function. The decoder follows a similar architecture with an increased hidden dimension of $1536$ with only 2 layers. Contrastive pretraining uses an additional projector with hidden dimension of $4096$ and output size of $256$ and operates on the concatenation of all class tokens as an input.

Pretraining is accomplished with a batch size of $4096$ for a total of $200M$ samples using the AdamW~\cite{Loshchilov2017FixingWD} optimizer with momentum parameters of $(0.9, 0.95)$ and weight decay of $5\times10^{-2}$. Weight decay is excluded from all learnable norms and biases. The base learning rate is  $1.5\times10^{-4}$ following a cosine schedule with linear warm-up and linear scaling rule with respect to a base batch size of $256$. For contrastive runs, the temperature is $0.2$. All runs are trained on Nvidia V100 GPUs with $\textrm{fp}16$ precision. Geometric augmentations are applied through random masking as well as consistent random rotations and flips applied along spatial axes for all experimental variations. Additionally, subsampling from larger quantized contexts to simulate random translations is applied in the contrastive setting. While rotations and flips do not alter embedding context, they aim to regularize learned position encodings.

\subsection{Evaluation models and metrics}
Baseline comparison models include AB-MIL models trained using embeddings from a single magnification, as well as the concatenation of all embeddings along the sequence dimension (referred to as all$\times$). Transformers  with the same architecture as the mixing module but with randomly initialized weights (i.e. no pretraining) are also trained jointly with an AB-MIL aggregation model and evaluated using the class token and contextualized embedding tokens.  All models are trained and tuned with label smoothing~\cite{7780677}.

Models are fine-tuned using group to slide label propagation as per equation \ref{eqn:s2g}. Both contextualized and compressed region embeddings are used to investigate the utility of the model across modes of operation. The region encoder and aggregator with a hidden dimension of $16$ are tuned with the AdamW~\cite{Loshchilov2017FixingWD} optimizer with a learning rate of $1\times10^{-5}$ and weight decay of $5\times10^{-2}$ with a batch size of 32. As per the pretraining recipe, weight decay is excluded from norm and bias parameters. The embedding encoder is first frozen for a single epoch and then trained jointly with the aggregator using a layer-wise learning rate decay~\cite{clark2020electrapretrainingtextencoders,Bao2021BEiTBP, he2022masked} of 0.75. Area under the receiver operator curve (AUROC) is used to evaluate the fine-tuned models and is reported on held-out test sets. 

\subsection{Evaluation tasks and data}
We evaluate performance of biomarker prediction across tissue types. As noted above, biomarker tasks are complex and have uncertain predictive features. Biomarker quantification is performed for seven biomarkers as measured by MSK-IMPACT~\cite{cheng2015-msk-impact}, a targeted genetic mutation test. The presence of biomarkers impacts therapeutic response and survival prediction. The specific biomarkers used in analysis are: Breast-CDH1, Colon-MSI, Bladder-FGFR, Endometrial PTEN, Lung-EGFR, Prostate-AR, Gastric-HER2, and Skin-BRAF. The training dataset sizes range from approximately 500 to 4500 (see Appendix Table A1 for complete details). For additional information regarding the biological implications of the biomarkers see~\cite{vorontsov2024virchow,shaikovski2024prism}.

\section{Results}
The results using the contextualized region embeddings and compressed region embeddings are presented in Table~\ref{tab:main_patch} and Table~\ref{tab:main_cls}, respectively. In all cases, on average, pretraining improves over baseline models and randomly initialized models. The impact of pretraining varies across tissue and biomarker type. Note that we refer to the contextualized region embeddings as Patch and the compressed region embedding as CLS as to indicate where the outputs are taken from the standard vision transformer model.

Notably, there are no settings in which the standard AB-MIL approach has the best performance, even when using all available magnifications. In all cases on average, we observe the randomly initialized mixing encoder improves upon AB-MIL for a given magnification, although the gains are modest. There are two biomarkers, endometrial PTEN and skin BRAF, where the randomly initialized mixing encoder has the best performance, although this result is not observed for a consistent input magnification.

\begin{table*}
\centering
\small
\begin{tabular}{llcccccccc}
\toprule
       &       & Breast & Bladder & Colon & Endo & Gastric & Lung & Skin & \multirow{2}{*}{Average}  \\ 
       &       & CDH1   & FGFR    & MSI   & PTEN & HER2    & EGFR & BRAF  & \\ \hline
\multirow{4}{*}{AB-MIL}    & 5$\times$    & 94.7        & 82.8    & 96.6      & 81.9 & 79.5         & 79.2 & 86.5      & 85.9     \\
       & 10$\times$   & 95.7        & 84.4    & 97.3      & 81.6 & 82.5         & 80.0 & 63.4      & 83.6     \\
       & 20$\times$   & 93.1        & 83.0    & 97.8      & 83.6 & 82.3         & 79.2 & 85.5      & 86.3     \\
       & all$\times$  & 90.9        & 84.7    & 98.0      & 82.1 & 83.1         & 78.6 & 88.9      & 86.6     \\ \hline
\multirow{4}{*}{Random}   & 5$\times$    & 94.3        & 86.9    & 95.8      & 85.0 & 78.2         & 76.8 & 85.3      & 86.0     \\
       & 10$\times$   & 91.9        & 86.6    & 96.1      & \textbf{86.3} & 74.9         & 78.3 & 89.1      & 86.2     \\
       & 20$\times$   & \underline{97.3}        & 86.3    & 96.8      & 85.8 & 66.5         & 78.5 & \textbf{95.8}      & 86.7     \\
       & all$\times$  & 96.6        & 81.5    & 97.8      & 85.4 & 75.0         & 78.4 & 94.1      & 87.0     \\ \hline
\multirow{3}{*}{MEM}      & 0.25  & \textbf{97.4}        & \underline{88.9}    & \underline{98.2}      & 85.9 & 81.7         & \textbf{84.4} & 94.5      & 90.1     \\
       & 0.50   & 96.1        & 88.3    & \textbf{98.7}      & \underline{86.0} & \textbf{86.9}         & 82.1 & 93.3      & \underline{90.2}     \\
       & 0.75  & \textbf{97.4}        & 88.8    & \underline{98.2}      & 85.5 & 82.3         & 81.6 & 93.9      & 89.7     \\ \hline
\multirow{3}{*}{CMEM}     & 3$\times$3   & 97.0        & 88.6    & 97.7      & 84.7 & 82.6         & \underline{83.7} & 93.7      & 89.7     \\
       & 7$\times$7   & 96.8        & \textbf{89.3}    & \underline{98.2}      & 85.9 & \underline{86.7}         & 81.4 & \underline{95.6}      & \textbf{90.5}     \\
       & 11$\times$11 & 97.0        & 88.6    & \underline{98.2}      & \underline{86.0} & 79.0         & 81.2 & \underline{95.6}      & 89.4    \\
       \bottomrule
\end{tabular}
\caption{Results measured by AUROC using the patch tokens, i.e., uncompressed representations, of the various models. The best and second best performance for each biomarker are bolded and underlined, respectively. On average, pretrained models outperform all baselines.}
\label{tab:main_patch}
\end{table*}
 
\begin{table*}[]
\centering
\small
\begin{tabular}{llcccccccc}
\toprule
       &       & Breast & Bladder & Colon & Endo & Gastric & Lung & Skin & \multirow{2}{*}{Average}  \\ 
       &       & CDH1   & FGFR    & MSI   & PTEN & HER2    & EGFR & BRAF  & \\ \hline

\multirow{4}{*}{Random} & 5$\times$    & 89.1                 & 86.5                 & 95.5                 & \textbf{86.5}        & 78.7                 & 76.3                 & 85.5                 & 85.4                 \\
                        & 10$\times$   & 93.1                 & 83.4                 & 96.5                 & 84.2                 & 78.5                 & 78.3                 & 91.5                 & 86.5                 \\
                        & 20$\times$   & 95.6                 & 85.5                 & 97.5                 & 84.2                 & 80.5                 & 77.5                 & 94.7                 & 87.9                 \\
                        & all$\times$  & \textbf{96.1}        & 82.9                 & 95.3                 & 83.8                 & 80.6                 & 77.9                 & \textbf{95.2}        & 87.4                 \\ \hline
\multirow{3}{*}{MEM}    & 0.25  & 94.7                 & 87.9                 & \underline{98.2}           & 85.4                 & \textbf{85.8}        & \textbf{83.5}        & \textbf{95.2}        & \textbf{90.1}        \\
                        & 0.50   & 95.5                 & \textbf{89.5}        & 96.5                 & 84.9                 & \underline{84.8}           & \underline{82.6}           & 89.9                 & \underline{89.1}           \\
                        & 0.75  & \underline{95.8}           & \underline{88.2}           & 97.9                 & \underline{85.7}           & 77.0                 & 79.6                 & \underline{94.3}           & 88.4                 \\  \hline
\multirow{3}{*}{CMEM}   & 3$\times$3   & 72.9                 & 77.9                 & 97.0                 & 77.9                 & 61.9                 & 75.0                 & 86.7                 & 78.5                 \\
                        & 7$\times$7   & 95.0                 & 83.4                 & \textbf{98.3}        & 85.4                 & 76.5                 & 80.1                 & 83.2                 & 86.0                 \\
                        & 11$\times$11 & 91.8                 & 83.7                 & 97.2                 & \underline{85.7}           & 74.7                 & 80.4                 & 89.5                 & 86.1                 \\ \bottomrule
\end{tabular}
\caption{Results measured by AUROC using the CLS tokens, i.e. compressed representations, of the various models. The best and second best performance for each biomarker are bolded and underlined, respectively. On average, MEM models perform the best. Notably, MEM performance using the CLS token has only about 1.0 decrease in AUROC compared to the uncompressed patch token performance.}
\label{tab:main_cls}
\end{table*}

\begin{figure*}
    \centering
    \includegraphics[trim={0 0 0 0},clip, width=0.95\linewidth]{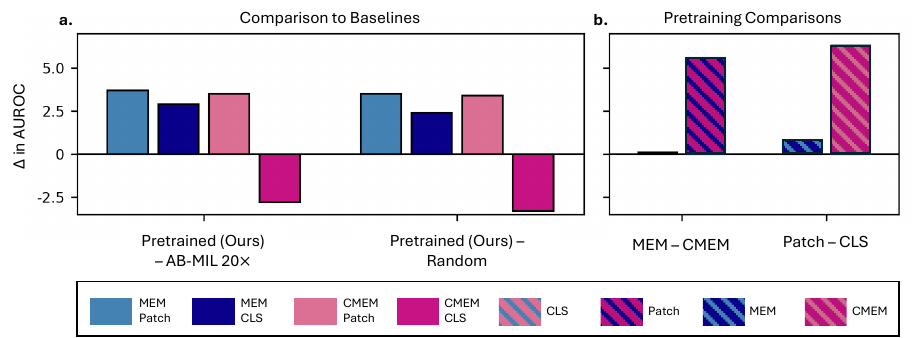}
    \caption{Summarized results of various aspects of the pretraining framework as measured by difference in AUROC (higher is better). ``Pretrained'' refers to MEM and CMEM results averaged over biomarker tasks and removal ratio $r$ and source size $c$, respectively. Random refers to the setting with no pretraining, averaged over possible magnifications. Overall, we observe improvements by using pretraining whether compared to AB-MIL or a randomly initialized model. MEM generally outperforms CMEM and the contextualized region (Patch) embeddings outperformed the compressed (CLS) embedding.}
    \label{fig:results_summary}
\end{figure*}

We do not observe a single setting to be the consistent best, however in most settings, MEM at any removal ratio outperforms alternatives. To summarize insights from the experimentation, we consider the impact of various design decisions on performance in aggregate (see Fig.~\ref{fig:results_summary}). Moving from left to right, we first observe that MEM pretraining averaged over removal ratio improves results across the board as compared to the literature standard of AB-MIL 20$\times$. CMEM averaged over source context sizes has less consistent performance, which is primarily degraded when using the compressed class token. These insights are also observed when comparing to mixing without pretraining (Pretrained v. Random). MEM slightly outperforms CMEM on average in the case of contextualized tile embedding, however CMEM with 7$\times$ input size has the best average overall. Surprisingly, the compressed region embedding of the MEM models outperforms the CMEM pretraining despite the additional supervisory signal on the CLS tokens. The decrease in performance from contextualized tile embeddings to compressed embedding for MEM models is relatively small.

Of the settings we have tested, our overall recommendation is for MEM pretraining with 50\% removal ratio. This results in an average improvement in AUROC of 3.9 as compared to AB-MIL at 20$\times$. The average improvements in AUROC compared to all$\times$ mixing without pretraining (random) for biomarker prediction is 3.2.

\section{Discussion}

In this work, we aim to challenge standard two-stage tile training pipelines by better leveraging the information contained within tile representations across multiple magnifications. We demonstrate the value of region-level representations, which are learned using only self-supervised signal, for both improving performance and decreasing sequence length.  Our results also support existing observations in the literature~\cite{campanella2023computational}, which suggest that 20$\times$ representations may not be optimal for all tasks. This further motivates the utility of leveraging multiple magnifications, particularly as the scope of CPath models continues to grow.

Few works have considered a similar approach. The vision-only version of the Titan model~\cite{ding2024multimodalslidefoundationmodel} is likely the most similar. Titan uses iBOT~\cite{Zhou2021iBOTIB} pretraining applied to grids of embeddings generated from a fixed magnification. 
The PRISM model~\cite{shaikovski2024prism}, a whole-slide-level representation task using pathology report data for training, also reports AUROC on the same biomarker prediction tasks. Although the studies are quite different and overly detailed comparisons may not be appropriate, we find it noteworthy that using the mixing encoder improved in all cases except breast-CDH1. CDH1 is strongly correlated with lobular cancer, which has a distinct histologic phenotype. In the PRISM study, CDH1 prediction most clearly benefited from language pretraining, perhaps suggesting that lobular cancer is often described in report and suggesting that for tasks without clear morphologic features, magnification mixing can provide an improvement not achievable with text.

Our work is unique in its aim of studying mixing magnifications. We do not necessarily claim to create the single best model but instead advocate for the addition of region-level representations to improve the performance of computational pathology models. This approach can be used with any multi-magnification foundation model, and we aimed to control for possible confounding of the foundation model by using only Virchow2 embeddings. In our study design, autoencoding was selected over latent reconstruction techniques such as iBOT, under the prior that foundation model features are already salient and do not require significant amount of transformation. Moreover, while reconstruction based tasks have been applied in the computational pathology domain on embeddings~\cite{Xu2024}, there is uncertainty on its efficacy, as reconstruction objectives without fine-tuning may not benefit perception based tasks~\cite{Balestriero2024LearningBR}. Based on our findings in conjunction with self-supervised learning theory \cite{vanassel2025jointembeddingvsreconstruction}, the types of features available in foundation model embedding space are subtle, thus preferring reconstruction based learning on average. 

Learning from images tiles and embeddings can be quite challenging, since the design space, which includes tile size, magnification, and algorithm is decided ad-hoc, and is validated based primarily on performance of diagnostic tasks such as pan-cancer detection. While the choice of region subsampling in a contrastive setting doubles down on how image foundation models are trained, it is clear that the learning signal extracts the wrong feature set for biomarker prediction tasks. Interestingly, the model naturally learns useful tokens in the MEM setting, where the compressed class token acts as a buffer.  

Somewhat surprisingly, compressed region embeddings still largely outperform baseline approaches, albeit by a smaller margin than contextualized region embeddings. The ability of the model to compress sequence lengths further motivates its utility in integrating with CPath models and suggests that it would be possible to further combine them into more complex vision-language systems by endowing them with the ability to reason across scales without overly large sequence lengths. Contrastive methods rely heavily on the design of augmentations which do not have natural analogues in embedding space beyond masking and cropping. The implementation and design of random crops over a large spatial region is ambiguous: the difficulty of the task is related to the source and target region sizes however, due to the combinatorial design space and uncertainty in defining meaningful learning signal, selecting source-target region pairs is challenging.

Overall, we recommend that pretrained mixed magnification representations be used in computational pathology workflows as they improve performance and can decrease sequence length without sacrificing accuracy. Importantly, the addition of these models does not preclude further aggregation into WSI representation, vision-language models, and multi-modal systems. Furthermore, they avoid assumptions on appropriate magnification for representations enabling flexibility for the many aims of CPath models.

\normalsize
\bibliography{main}

\footnotesize
\clearpage
\onecolumn
\setcounter{page}{1}

\section*{Supplementary Material: Training Data}

\begin{table}[H]
\centering
\begin{tabular}{@{}rrrrr@{}}
\toprule
Biomarker & Subset & Cases & Slides & Positive \\ \midrule
\multirow{3}{*}{Breast-CDH1}      & train  & 648   & 673    & 0.15     \\
                                  & tune   & 214   & 219    & 0.14     \\
                                  & test   & 214   & 228    & 0.15     \\\midrule
\multirow{3}{*}{Colon-MSI}        & train  & 2008  & 2268   & 0.10      \\
                                  & tune   & 335   & 381    & 0.12     \\
                                  & test   & 331   & 373    & 0.13     \\\midrule
\multirow{3}{*}{Bladder-FGFR}     & train  & 518   & 540    & 0.26     \\
                                  & tune   & 255   & 271    & 0.31     \\
                                  & test   & 254   & 265    & 0.28     \\\midrule
\multirow{3}{*}{Endometrial-PTEN} & train  & 975   & 1027   & 0.48     \\
                                  & tune   & 162   & 168    & 0.43     \\
                                  & test   & 164   & 178    & 0.41     \\ \midrule
\multirow{3}{*}{Lung-EGFR}        & train  & 2129  & 2773   & 0.28     \\
                                  & tune   & 345   & 437    & 0.30     \\
                                  & test   & 352   & 446    & 0.28     \\\midrule
\multirow{3}{*}{Gastric-Her2}     & train  & 967   & 967    & 0.19     \\
                                  & tune   & 169   & 169    & 0.23     \\
                                  & test   & 160   & 160    & 0.18     \\ \midrule
\multirow{3}{*}{Skin-BRAF}        & train  & 320   & 343    & 0.23     \\
                                  & tune   & 55    & 58     & 0.11     \\
                                  & test   & 54    & 62     & 0.13     \\
                                  \bottomrule
\end{tabular}
\caption{A summary of the data used for the biomarker prediction tasks. For each biomarker the number of cases, slides (whole slide images) and positivity rate for the train, tune, and test splits is reported.}
\label{tab:supp_biomarker-data-stats}
\end{table}

\end{document}